\documentclass[runningheads,a4paper]{llncs}

\usepackage{amssymb}
\usepackage{graphicx}
\usepackage{multirow}
\usepackage{amsmath}
\usepackage{stfloats}
\usepackage{amsfonts}

\usepackage{url}
\urldef{\mailsa}\path|{alfred.hofmann, ursula.barth, ingrid.haas, frank.holzwarth,|
\urldef{\mailsb}\path|anna.kramer, leonie.kunz, christine.reiss, nicole.sator,|
\urldef{\mailsc}\path|erika.siebert-cole, peter.strasser, lncs}@springer.com|
\newcommand{\keywords}[1]{\par\addvspace\baselineskip
\noindent\keywordname\enspace\ignorespaces#1}

\begin{document}

\mainmatter  

\title{Generative Model Watermarking Based on Human Visual System}

\titlerunning{Generative Model Watermarking Based on Human Visual System}

%
%
\author{Li Zhang$^{1,\dagger}$, Yong Liu$^{1,\dagger}$, Shaoteng Liu$^2$, Tianshu Yang$^2$, Yexin Wang$^3$, Xinpeng Zhang$^1$ and Hanzhou Wu$^{1,\star}$}
\authorrunning{L. Zhang, Y. Liu, S. Liu \emph{et al.}}
\institute{
$^1$School of Communication and Information Engineering, Shanghai University, Shanghai 200444, Shanghai, China\\
$^2$AI Technology Center, Tencent Inc., Beijing 100080, Beijing, China\\
$^3$AI Technology Center, Tencent Inc., Shenzhen 518000, Guangdong, China\\
jmzhangli@shu.edu.cn, liuyongresearch@163.com, \{shaotengliu, gracetsyang, yexinwang\}@tencent.com, xzhang@shu.edu.cn, h.wu.phd@ieee.org\\
$^\dagger$Equally contributed authors, $^\star$Corresponding author}
%
%

\maketitle

\begin{abstract}
Intellectual property protection of deep neural networks is receiving attention from more and more researchers, and the latest research applies model watermarking to generative models for image processing. However, the existing watermarking methods designed for generative models do not take into account the effects of different channels of sample images on watermarking. As a result, the watermarking performance is still limited. To tackle this problem, in this paper, we first analyze the effects of embedding watermark information on different channels. Then, based on the characteristics of human visual system (HVS), we introduce two HVS-based generative model watermarking methods, which are realized in RGB color space and YUV color space respectively. In RGB color space, the watermark is embedded into the R and B channels based on the fact that HVS is more sensitive to G channel. In YUV color space, the watermark is embedded into the DCT domain of U and V channels based on the fact that HVS is more sensitive to brightness changes. Experimental results demonstrate the effectiveness of the proposed work, which improves the fidelity of the model to be protected and has good universality compared with previous methods.

\end{abstract}

\section{Introduction}
Deep learning (DL) \cite{DLNature} has been applied in many application fields such as computer vision \cite{Unet} and natural language processing \cite{NLP}. It is foreseen that DL will continue bringing profound changes to our daily life. However, it is known that training a powerful DL model requires lots of computational resources and well-labelled data. As a result, the trained DL model should be regarded as an important digital asset and should be protected from intellectual property (IP) infringement. Therefore, how to protect the IP of DL models has become a new technical challenge. It motivates researchers and engineers from both academia and industry to exploit \emph{digital watermarking} \cite{cox:book} for IP protection of DL models \cite{Uchida2017}, which is typically referred to as \emph{model watermarking}. However, unlike conventional media watermarking that often treats the host signal as static data, DL models possess the ability of accomplishing a particular task, indicating that the algorithmic design of model watermarking should take into account the functionality of DL models.

An increasing number of model watermarking schemes are proposed in recent years. By reviewing the mainstream algorithms, advanced strategies realizing model watermarking can be roughly divided into three main categories, i.e., network based model watermarking, trigger based model watermarking, and dataset based model watermarking. Network based model watermarking algorithms aim at covertly embedding a watermark into the internal network weights \cite{Uchida2017}, \cite{Li2021}, \cite{WangSPIE2020} or network structures \cite{Zhao2021WIFS}. As a result, the watermark decoder should access the internal details fully or partly of the target DL model. Though trigger based model watermarking algorithms often change the internal parameters of the DL model through training from scratch or fine-tuning, the zero-bit watermark is carried by the mapping relationship between a set of trigger samples and their pre-specified labels. As a result, the watermark is extracted by analyzing the consistency between the labels and the prediction results of the target DL given a set of trigger samples \cite{adi:2018}, \cite{Zhao2021ISDFS}. Dataset based model watermarking algorithms exploit the entangled relationship between the DL model and a dataset related to the task of the DL model to construct a fingerprint of the DL model, which may be realized via a lossless way (i.e., without modifying the original model to be protected) to guarantee high security such as \cite{Liu:TIFS}, \cite{Lin:ISDFS}, \cite{Wu:PMI}.

Recently, Wu \emph{et al.} \cite{wuTCSVT2021} propose a general watermarking framework for deep generative models. By optimizing a combined loss function, the trained model not only performs very well on its original task, but also automatically inserts a watermark into any image outputted by the model. In the work, the watermark is embedded into the model through watermark-loss back-propagation. As a result, by extracting the watermark from the outputted image, the ownership of the target model can be identified without interacting directly with the model. Yet another similar work is introduced by Zhang \emph{et al.} \cite{Zhangjie:2020}, which was originally designed to resist surrogate attacks and also allows us to extract a watermark from the outputted image. Though \cite{wuTCSVT2021} and \cite{Zhangjie:2020} have demonstrated satisfactory watermark verification performance as reported in the papers, they both do not consider the characteristics of the human visual system (HVS). As a result, the visual quality may not be satisfactory from a broader perspective.

To tackle the aforementioned problem, we utilize the characteristics of HVS for system design of model watermarking. To this end, we propose two HVS based frameworks for generative model watermarking, i.e, HVS-RGB based generative model watermarking and HVS-YUV based generative model watermarking. Since the proposed two frameworks take into account the HVS characteristics, the generated \emph{marked} images are more compatible with the HVS, thereby improving the fidelity of the network and the invisibility of the watermark. In summary, the main contributions of this paper include:

\begin{itemize}
  \item We have analyzed the impact of embedding watermark on different channels of the generated images in the generative model watermarking by combining the HVS characteristics. Based on this analysis, we propose two HVS-based multi-channel generative model watermarking frameworks that improve the fidelity of the network while maintaining the fidelity of the watermark.
  \item Through convinced experiments on semantic segmentation, the results demonstrate the applicability and superiority of the proposed work.
\end{itemize}

The rest structure of this paper is organized as follows. In Section 2, we give the details of the proposed work. Then, we provide experiments and analysis in Section 3. Finally, we conclude this paper and provide discussion in Section 4.

\begin{figure*}[!t]
  \centering
  \includegraphics[width=\linewidth]{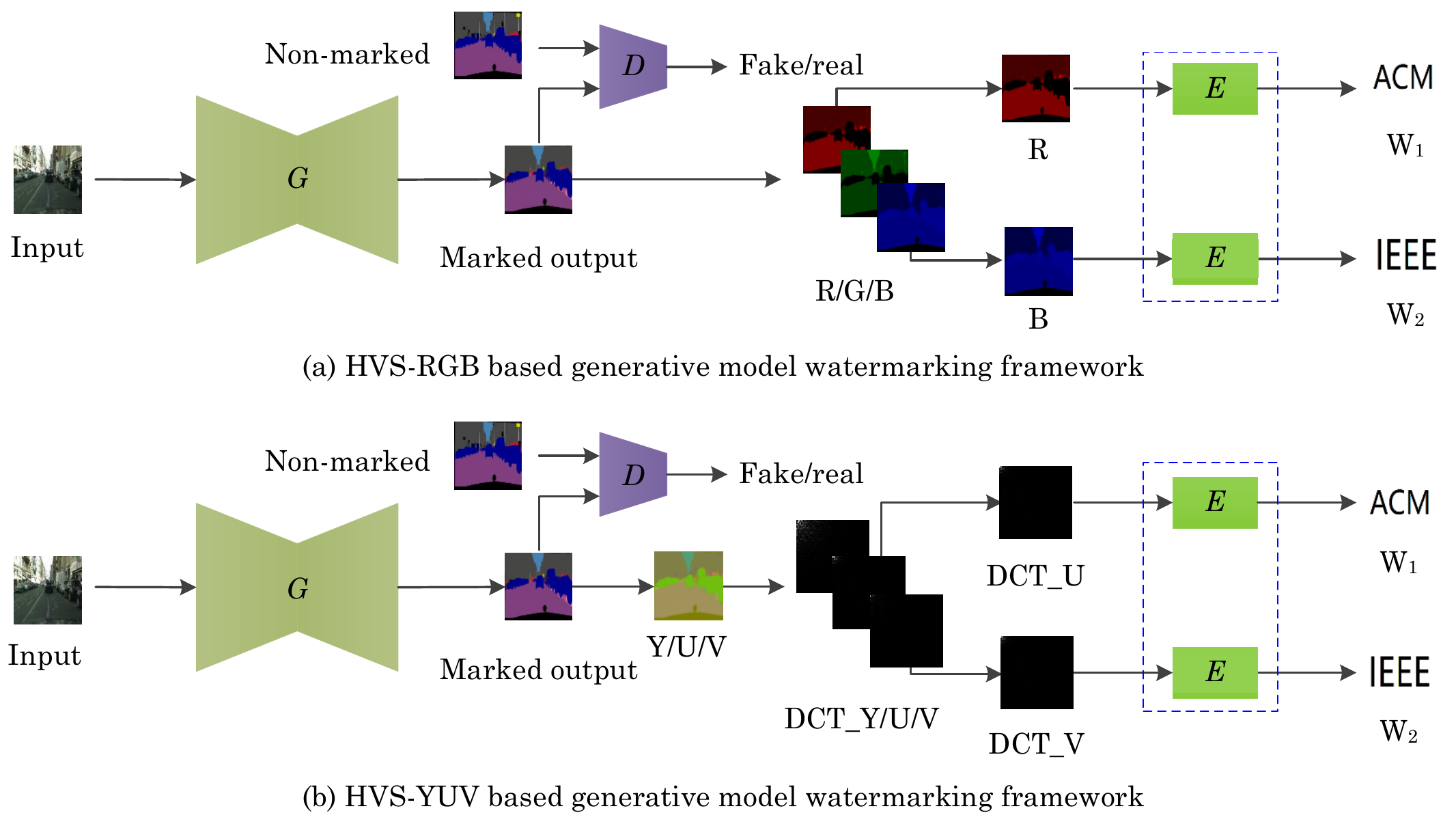}\\
  \caption{The proposed two HVS based frameworks for generative model watermarking. For better understanding, the original task of $G$ here is set to semantic segmentation.}\label{F1}
\end{figure*}

\section{Proposed Method}

\subsection{HVS Characteristics}
HVS is one of the most complex biological systems \cite{HVS1:paper}, \cite{HVS2:paper}, \cite{HVS3:paper} and typically includes the following characteristics:

\begin{itemize}
  \item \emph{Spectral characteristics:} The sensitivity of HVS to color is green, red, blue in descending order. Therefore, embedding a watermark in the green channel of a color image will affect the visual quality of the image more than the others. It is desirable to preferentially embed the watermark into the blue channel and then the red channel. The green component can be changed with the least degree in order to improve the invisibility of the watermark.
  \item \emph{Brightness characteristics:} The HVS is more sensitive to changes in brightness. Therefore, in YUV color space, embedding the watermark into U or V chromatic aberration channel results in better visual quality than Y channel.
\end{itemize}

The brightness and spectral characteristics of HVS illustrate the effects of embedding the watermark into different channels of the image. In conventional image watermarking, there are already methods verifying that HVS characteristics are helpful for developing high fidelity watermarking systems such as \cite{HVS3:paper}, \cite{HVS4:paper}. In this paper, we mainly combine the spectral and brightness characteristics to analyze the effects of watermark embedding in different channels of the generated image in generative model watermarking. In the later experiments, we will analyze the visual quality of the generated images after embedding the watermark in different channels, and analyze the results of the experiments by combining the spectral and brightness characteristics of HVS. It is proved that the HVS characteristics still hold in the generative model watermarking. Therefore, the HVS characteristics can guide us to design model watermarking methods that better match the characteristics of human visual perception system.

\subsection{HVS-based Generative Model Watermarking Frameworks}
We propose two HVS-based generative model watermarking (GMW) frameworks, which are shown in Figure \ref{F1}. Below, we first provide the overview and then describe the watermark embedding and extraction process.

\textbf{HVS-RGB based GMW framework:} As mentioned above, human beings are more sensitive to the green color. Reducing the modifications to the green (G) channel results in better visual quality of the image, which inspires us to preferentially embed the watermark information into the red (R) channel and/or the blue (B) channel. Motivated by this insight, we propose a framework to embed two watermarks into the R channel and the B channel respectively. As shown in Figure \ref{F1}, the process can be described as follows. Given the generative model $G$ to be protected, we train $G$ in such a way that the image outputted by $G$ is always marked. The watermark information is carried by the R channel and the B channel of the marked image. Specifically, we determine the R/G/B channels of the image outputted by $G$. The R channel of the image is used to carry a watermark $W_1$ and the B channel is used to carry another watermark $W_2$. To realize this purpose, a neural network $E$ is trained together with $G$ through loss optimization so that $W_1$ can be extracted from the R channel, and $W_2$ can be extracted from the B channel. To ensure that $G$ has good performance on its original task, the image outputted by $G$ may be fed to a discriminator $D$. Though we consider $W_1$ and $W_2$ as two different watermarks in this paper, one may consider them as a whole, which will not cause any problem.

\textbf{HVS-YUV based GMW framework:} Similarly, human beings are more sensitive to changes in brightness. Therefore, by representing an image in the YUV color space, it is suggested to embed the watermark information in the U/V channels to not arouse noticeable artifacts. However, directly embedding watermark information into the U/V channels in the spatial domain may lead to significant distortion to the chroma. To address this problem, we propose to embed the watermark information into the U/V channels in the discrete cosine transform (DCT) domain. In this way, the entire process is similar to the above HVS-RGB based framework. The only one difference is that the image outputted by $G$ should be converted to the Y/U/V color space. Then, the U channel and the V channel are processed by DCT. The transformed data can be expressed as DCT\_U and DCT\_V. Thereafter, DCT\_U and DCT\_V are fed to $E$ for watermark embedding and watermark extraction.

Both frameworks are similar to each other according to the above analysis. For the two frameworks, we are now ready to describe the watermark embedding and extraction procedure for the host network $G$ as follows:

\begin{itemize}
  \item \emph{Watermark embedding:} The watermark information is embedded into $G$ by marking the output of $G$ through optimizing a combined loss that will be detailed later. In other words, $G$ is marked during joint training. The networks to be jointly trained include $G$, $E$ and $D$, where $D$ is optional.
  \item \emph{Watermark extraction:} After training, $G$ is deemed marked. By retrieving the watermark information from any image outputted by $G$ with $E$, we are able to trace the source of the image and identify the ownership of the image and the target model. As multiple watermarks can be embedded, multiple purposes can be achieved.
\end{itemize}

\begin{figure*}[!t]
  \centering
  \includegraphics[width=\linewidth]{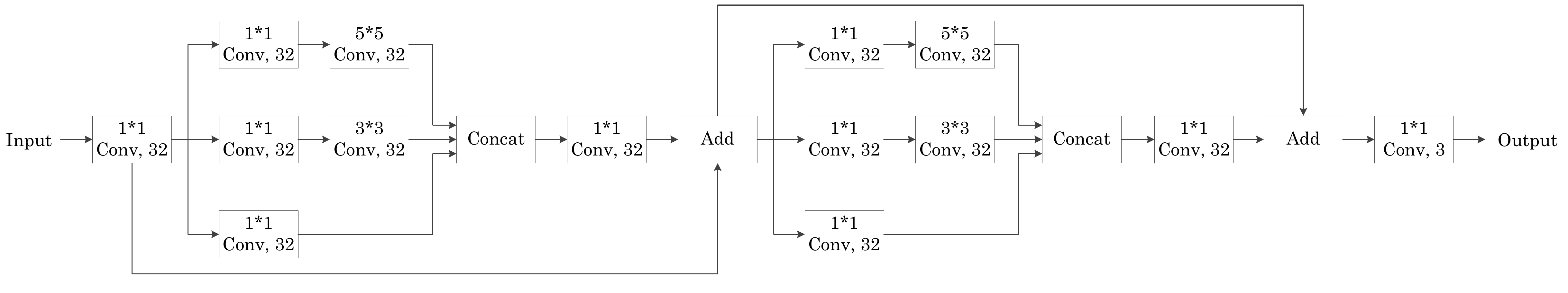}\\
  \caption{Structural information for the watermark-extraction network $E$.}\label{F2}
\end{figure*}

\subsection{Structural Design}
$G$ can perform any task outputting color images as results. In this paper, without the loss of generalization, we use the backbone network introduced in \cite{HDH:paper} as the host network $G$ to be protected. The network structure is inspired by U-net \cite{Unet}. The original task of $G$ is limited to image semantic segmentation, whose goal is to cluster parts of an image together which belong to the same object class. The input and the output have the same size in this paper.

The objective of $E$ is to extract the corresponding watermark from the corresponding channel (involving R, B, DCT\_U, DCT\_V). $E$ accepts a single channel as input and then outputs the corresponding watermark (that may be either a binary/gray-scale image or a color one). It is free to design the network structure of $E$. In this paper, the structure of $E$ is inspired by \cite{wuTCSVT2021}, \cite{EF:structure}. ReLU \cite{ReLU:paper} is applied as the activation function for the internal layers, except that the output layer uses TanH. The details of $E$ are shown in Figure \ref{F2}. In addition, $G$ can be optionally optimized by a discriminator $D$. For example, we may use PatchGAN \cite{PatchGAN:paper} to serve as $D$ shown in Figure \ref{F1}.

\subsection{Loss Function}
Based on the aforementioned analysis, we are now ready to design the loss function for each watermarking framework. We use $S_0$ to denote the set of images not generated by $G$ and $S_1$ to denote the set of images generated by $G$.

\textbf{Loss function for HVS-RGB based GMW framework:} For the watermarking task, we hope that $E$ can extract the corresponding watermark given the corresponding input, which can be expressed as:
\begin{equation}
L_1 = \frac{1}{|S_1|}\sum_{G(x)\in S_1}||E(G(x)_\text{R}) - W_1||
\end{equation}
and
\begin{equation}
L_2 = \frac{1}{|S_1|}\sum_{G(x)\in S_1}||E(G(x)_\text{B}) - W_2||
\end{equation}
where $x$ is the input image of $G$, $G(x)_\text{R}$ and $G(x)_\text{B}$ represent the R component and the B component of the outputted image $G(x)$, respectively. On the other hand, we hope that random noise can be extracted from images not generated by $G$, which requires us to minimize:
\begin{equation}
L_3 = \frac{1}{|S_0|}\sum_{y\in S_0}||E(y_\text{R}) - W_z||
\end{equation}
and
\begin{equation}
L_4 = \frac{1}{|S_0|}\sum_{y\in S_0}||E(y_\text{B}) - W_z||
\end{equation}
where $y_\text{R}$ and $y_\text{B}$ represent the R component and the B component of the image $y\in S_0$, and, $W_z$ is a noise image randomly generated in advance. For the original task of $G$, it is expected that the \emph{marked} image generated by $G$ is close to the ground-truth image. Typically, to accomplish the task, we expect to minimize:
\begin{equation}
L_5 = \frac{1}{|S_1|}\sum_{G(x)\in S_1}||G(x) - T[G(x)]||
\end{equation}
where $T[G(x)]$ means the ground-truth image of $G(x)$. Since the R component and the B component of $G(x)$ will be used to carry the additional information, distortion will be introduced. To reduce the impact of watermark embedding, it is desirable to further control the distortion in the R channel and the B channel, i.e., we hope to minimize:
\begin{equation}
L_6 = \frac{1}{|S_1|}\sum_{G(x)\in S_1}||G(x)_\text{R} - T_\text{R}[G(x)]||
\end{equation}
and
\begin{equation}
L_7 = \frac{1}{|S_1|}\sum_{G(x)\in S_1}||G(x)_\text{B} - T_\text{B}[G(x)]||
\end{equation}
where $T_\text{R}[G(x)]$ and $T_\text{B}[G(x)]$ represent the R component and the B component of $T[G(x)]$. Simply minimizing the above losses may not well maintain the structural information of the image, to deal with this problem, we further introduce structural loss for improving the image quality. Namely, we want to minimize:
\begin{equation}
L_8 = \frac{1}{|S_1|}\sum_{G(x)\in S_1}(1 - \frac{\text{SSIM}(G(x), T[G(x)]) + \text{MS-SSIM}(G(x), T[G(x)])}{2})
\end{equation}
where $\text{SSIM}(\cdot, \cdot)$ \cite{SSIM:paper} and $\text{MS-SSIM}(\cdot, \cdot)$ \cite{Wang2:paper} are two indicators representing the structural similarity between two images. The entire loss function is then:
\begin{equation}
L_\text{HVS-RGB} = \alpha(L_1+L_2)+\beta(L_3+L_4) + (L_5+L_6+L_7+L_8)
\end{equation}
where $\alpha$ and $\beta$ are pre-determined parameters balancing the losses.

\textbf{Loss function for HVS-YUV based GMW framework:} Similarly, the entire loss function can be divided into two parts, i.e., the watermarking loss and task loss. Referring to the loss function for HVS-RGB based GMW framework, the components of watermarking loss function include:
\begin{equation}
L_9 = \frac{1}{|S_1|}\sum_{G(x)\in S_1}||E(G(x)_{\text{DCT}\_\text{U}}) - W_1||
\end{equation}
\begin{equation}
L_{10} = \frac{1}{|S_1|}\sum_{G(x)\in S_1}||E(G(x)_{\text{DCT}\_\text{V}}) - W_2||
\end{equation}
\begin{equation}
L_{11} = \frac{1}{|S_0|}\sum_{y\in S_0}||E(y_{\text{DCT}\_\text{U}}) - W_z||
\end{equation}
\begin{equation}
L_{12} = \frac{1}{|S_0|}\sum_{y\in S_0}||E(y_{\text{DCT}\_\text{V}}) - W_z||
\end{equation}
where $G(x)_{\text{DCT}\_\text{U}}$ and $G(x)_{\text{DCT}\_\text{V}}$ represent the U channel and the V channel in the DCT domain for $G(x)$, respectively. $y_{\text{DCT}\_\text{U}}$ and $y_{\text{DCT}\_\text{V}}$ are the U channel and the V channel in the DCT domain for $y$. On the other hand, the task loss function involves $L_5$, $L_8$ and
\begin{equation}
L_{13} = \frac{1}{|S_1|}\sum_{G(x)\in S_1}||G(x)_{\text{DCT}\_\text{U}} - T_{\text{DCT}\_\text{U}}[G(x)]||
\end{equation}
\begin{equation}
L_{14} = \frac{1}{|S_1|}\sum_{G(x)\in S_1}||G(x)_{\text{DCT}\_\text{V}} - T_{\text{DCT}\_\text{V}}[G(x)]||
\end{equation}
where $T_{\text{DCT}\_\text{U}}[G(x)]$ and $T_{\text{DCT}\_\text{V}}[G(x)]$ represent the U component and the V component of $T[G(x)]$ in the DCT domain. So, the entire loss function is:
\begin{equation}
L_\text{HVS-YUV} = \alpha(L_9+L_{10})+\beta(L_{11}+L_{12}) + (L_5+L_8+L_{13}+L_{14}).
\end{equation}

\emph{Remark:} There are two choices for determining the network parameters of $G$ and $E$. One is to train $G$ and $E$ together from scratch according to the entire loss function. The other is to train $G$ and $E$ from scratch according to the task loss of $G$ and the watermark-extraction loss corresponding to $W_1$. Then, $G$ and $E$ can be fine-tuned with the entire loss function. We use the latter strategy due to the relatively lower computational complexity.

\begin{table}[!t]
\renewcommand{\arraystretch}{1}
\centering
\caption{Quality evaluation for the marked images generated by the host network.}
\begin{tabular}{c|c|c|c}
\hline\hline
Channel & Average PSNR (dB) & Average SSIM & Average MS-SSIM\\
\hline
R & 23.74 & 0.798 & 0.853\\
G & 22.92 &	0.782 & 0.842\\
B & 23.63 &	0.801 & 0.858\\
\hline
DCT\_Y & 22.85 & 0.781 & 0.840\\
DCT\_U & 23.49 & 0.794 & 0.848\\
DCT\_V & 23.60 & 0.793 & 0.853\\
\hline\hline
\end{tabular}\label{T1}
\end{table}

\section{Experimental Results and Analysis}
\subsection{Setup}
As mentioned previously, we limit the task of $G$ to image semantic segmentation. And, the structure of $G$ is similar to U-net. However, it should be admitted that it is always free to design the network structure of $G$ and specify the task of $G$. Since our purpose is not to develop $G$, it is true that the network structure used in this paper is not optimal in terms of the task performance, which, however, does not affect the contribution of this work. We use the dataset mentioned in \cite{ssdataset:paper} for experiments, where 4,000 images are used for model training, 500 images are used for validation and another 500 images are used for testing. The size of input and the size of output are set to $256\times 256\times 3$ for the host network. For $W_1$, $W_2$ and $W_z$, their sizes are all equal to $256\times 256\times 3$ since the watermark-extraction network in Figure \ref{F2} keeps the size of the output same as the size of the input, indicating that the size of the output should be identical to that of the image outputted by $G$ as the latter will be fed into the network in Figure \ref{F2}.

During model training, we empirically applied $\alpha = 1$ and $\beta = 0.5$. The Adam optimizer was used to iteratively update the network parameters. The learning rate was set to $2\times 10^{-4}$. Our implementation used TensorFlow trained on a single RTX 3090 GPU. We use three evaluation metrics, i.e., peak signal-to-noise ratio (PSNR), structural similarity (SSIM) \cite{SSIM:paper} and multiscale structural similarity (MS-SSIM) \cite{Wang2:paper} for evaluation. For all of these widely used indicators, a higher value demonstrates the better quality of the image. In this paper, we limit $W_1$ to the logo image ``ACM'' and $W_2$ to the logo image ``IEEE'' (refer to Figure \ref{F1}). Besides, $W_z$ is a randomly generated noise image. $\ell_1$ norm is applied to all the loss functions, which can speed up convergence during training.

\subsection{Analysis of Watermark Embedding in Different Channels}
Before reporting experimental results, we first analyze the effects of watermark embedding on different channels of the image by taking into account the aforementioned spectral and brightness characteristics. To this purpose, we separate three channels from the image generated by the host network and feed each of the channels into the watermark-extraction network. For each channel, the host network together with the watermark-extraction network are trained together so that the image outputted by the host network carries a watermark that can be extracted by the watermark-extraction network. It is pointed that we here only tested the watermark $W_1$, where $W_2$ was not used for simplicity. In other words, all the channels independently carry the same watermark for fair comparison.

Table \ref{T1} provides the experimental results of image quality evaluation. It can be found that even though different color channels result in different performance in terms of image quality, overall, R/B and DCT\_U/DCT\_V are superior to G and DCT\_Y, respectively. This indicates that from the viewpoint of watermark embedding, it is quite desirable to preferentially embed the watermark data into the R/B and DCT\_U/DCT\_V channels, rather than the G and DCT\_Y channels.

\begin{table}[!t]
\renewcommand{\arraystretch}{1}
\centering
\caption{Quality evaluation for the marked images and the extracted watermarks.}
\begin{tabular}{c|c|c|c|c|c}
\hline\hline
Method & Mean PSNR & Mean SSIM & Mean MS-SSIM & Mean BER$_1$ & Mean BER$_2$\\
\hline
HVS-RGB & 23.12 & 0.807 & 0.857 & 0.0024 & 0.0029\\
HVS-YUV & 22.83 & 0.799 & 0.852 & 0.0026 & 0.0031\\
\cite{wuTCSVT2021} & 20.83 & 0.738 & 0.816 & 0.0020 & 0.0028\\
\hline\hline
\end{tabular}\label{T2}
\end{table}

\begin{figure*}[!t]
  \centering
  \includegraphics[width=4.2in]{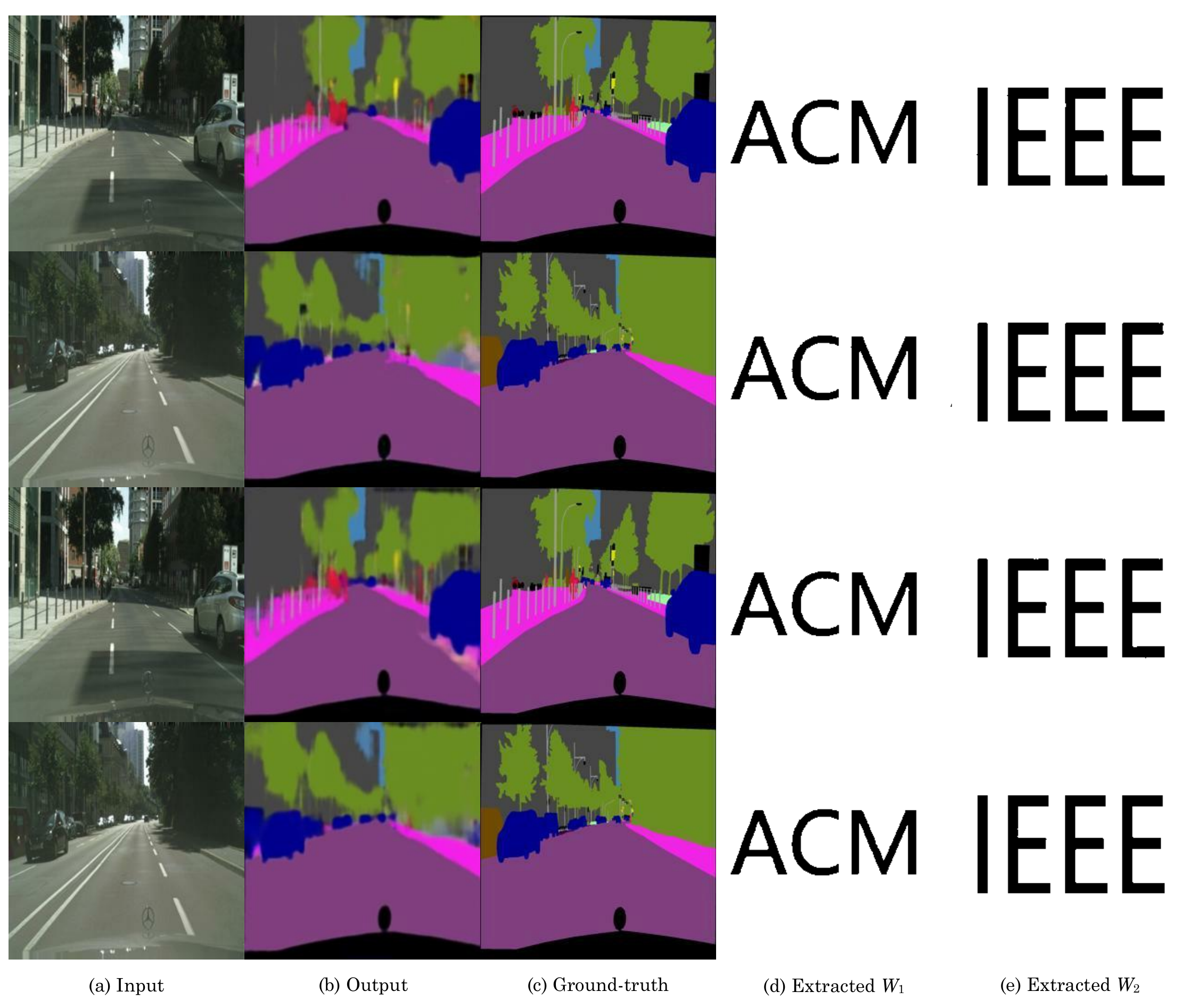}\\
  \caption{Visual examples for the marked images and the extracted watermarks. The first two rows correspond to the HVS-RGB framework and the last two rows correspond to the HVS-YUV framework. Notice that, the input in the first row is identical to that in the third row, and the input in the second row is identical to that in the fourth row.}\label{F3}
\end{figure*}

\subsection{Main Results}
The main contribution of this paper is to improve the visual quality of the marked images generated by the host network by taking into account the characteristics of HVS. Therefore, in this subsection, we have to measure the difference between the generated marked images and the ground-truth images. It is necessary that the embedded watermark information should be extracted with high fidelity for reliable ownership verification. Accordingly, the difference between the extracted watermark and the corresponding ground-truth should be analyzed as well.

We use PSNR, SSIM and MS-SSIM to quantify the visual difference between the images generated by the marked network and the ground-truth images. As shown in Figure \ref{F1}, two watermarks are embedded into the generated image. To quantify the watermarking performance, bit error rate (BER) is used for both $W_1$ and $W_2$. For fair comparison, we also realize a baseline watermarking system based on \cite{wuTCSVT2021} (without secret key), which feeds the image outputted by $G$ directly into the watermark-extraction network to reconstruct either $W_1$ or $W_2$.

Table \ref{T2} shows the experimental results, where ``mean BER$_1$'' is used for $W_1$ and ``mean BER$_2$'' is used for $W_2$. The others are used for the images generated by the marked neural network. It is observed from Table \ref{T2} that the BER difference between the proposed strategy and the strategy in \cite{wuTCSVT2021} is very low, which indicates that the proposed strategy does not impair the watermark fidelity. On the other hand, the proposed two frameworks significantly improve the quality of the image generated by the network. It means that the introduction of HVS is indeed helpful for enhancing the performance of generative model watermarking. Figure \ref{F3} further provides some examples, from which we can infer that the images are with satisfactory quality, which verifies the applicability.

\begin{figure*}[!t]
  \centering
  \includegraphics[width=\linewidth]{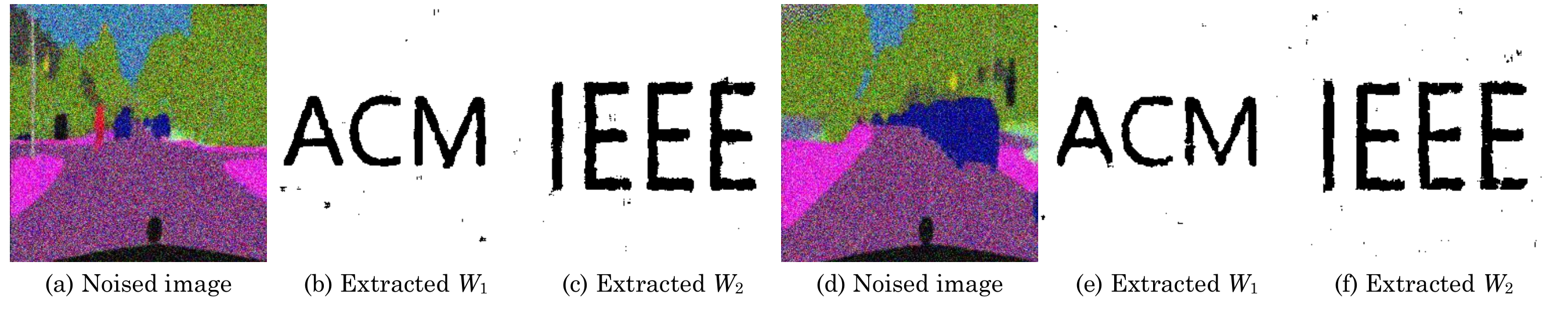}\\
  \caption{Visual examples for the noised images (marked) and the extracted watermarks. (a-c) HVS-RGB based framework and (d-f) HVS-YUV based framework. Here, $\sigma = 0.4$.}\label{F4}
\end{figure*}

\begin{figure*}[!t]
  \centering
  \includegraphics[width=4.2in]{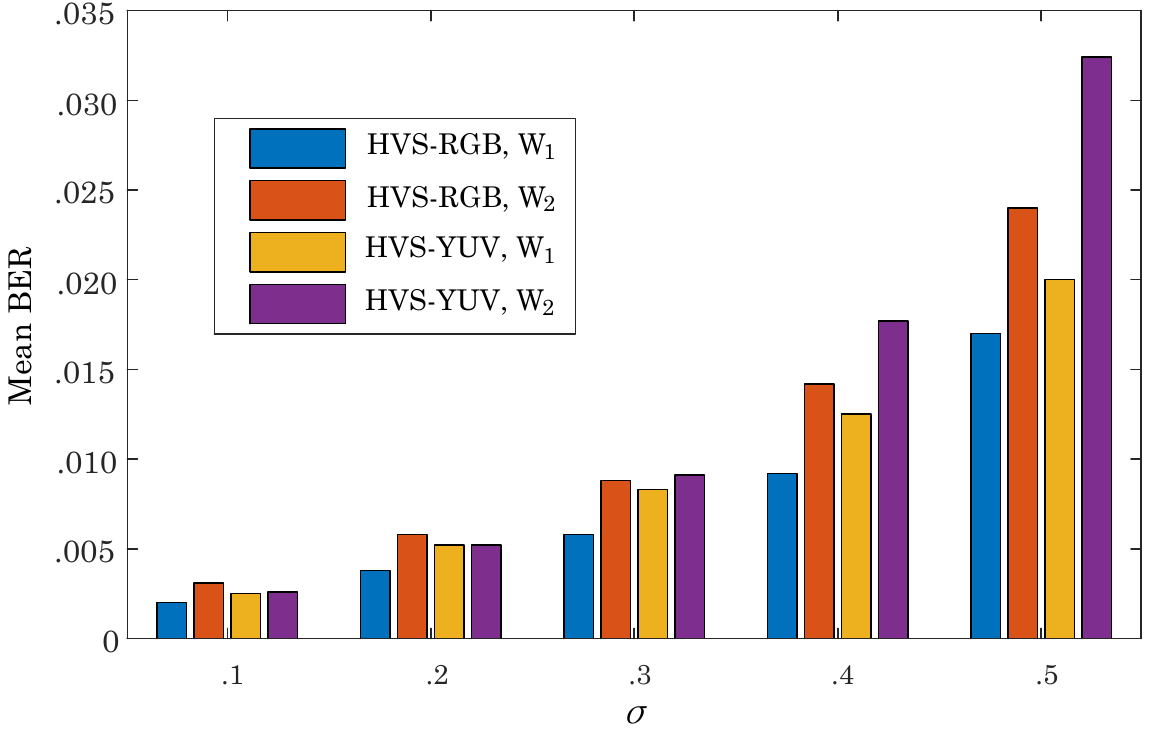}\\
  \caption{The mean BERs due to different degrees of noise addition.}\label{F5}
\end{figure*}

\subsection{Robustness and Ablation Study}
In application scenarios, the marked images generated by the host network may be attacked. It is necessary to evaluate the robustness of the proposed two frameworks against common attacks. To deal with this problem, it is suggested to mimic the real-world attack to the images generated by the host network during model training, which has been proven to be effective in improving the robustness of DL models. To this end, to evaluate the robustness, during model training, we mimic the real-world attack to the images generated by the host network and feed them into the watermark-extraction network for watermark extraction. We consider one of the most representative attacks, i.e., noise addition, for simplicity.

Specifically, during model training, we add the Gaussian noise to every image generated by the host network by applying $\mu = 0$ and random $\sigma \in (0, 0.5)$. The noised image is then fed into the watermark-extraction network for watermark extraction. As a result, the trained model has the ability to resist noise addition. In this way, we are able to evaluate the robustness of the trained model by using BER. Figure \ref{F4} provides an example of noise addition, from which we can infer that the visual quality of the marked image is significantly distorted, but the embedded watermark can be extracted with satisfactory quality. Figure \ref{F5} further provides the BERs due to different degrees of noise addition. It can be observed that the BER tends to increase as the degree of noise addition increases, which is reasonable because a larger degree of noise addition reduces the watermark information carried by the generated image, thereby resulting in a higher BER. However, overall, the BERs are in a low level. It indicates that by combining robustness enhancement strategies, the proposed two frameworks have satisfactory ability to resist against malicious attacks.

\begin{table}[!t]
\renewcommand{\arraystretch}{1}
\centering
\caption{Performance comparison by applying different loss functions. The experimental results in this table are mean values.}
\begin{tabular}{c|c|c|c|c|c|c|c}
\hline\hline
Method & SSIM Loss & MS-SSIM Loss & PSNR & SSIM & MS-SSIM & BER$_1$ & BER$_2$\\
\hline
HVS-RGB &            &            & 23.01 & 0.779 & 0.839 & 0.0029 & 0.0028\\
HVS-RGB & \checkmark &            & 23.05 & 0.801 & 0.846 & 0.0031 & 0.0030\\
HVS-RGB &            & \checkmark & 22.99 & 0.788 & 0.852 & 0.0026 & 0.0034\\
HVS-RGB & \checkmark & \checkmark & \textbf{23.12} & \textbf{0.807} & \textbf{0.857} & 0.0024 & 0.0029 \\
\hline
HVS-YUV &            &            & 22.57 & 0.778 & 0.835 & 0.0022 & 0.0025 \\
HVS-YUV & \checkmark &            & 22.79 & 0.784 & 0.838 & 0.0029 & 0.0034 \\
HVS-YUV &            & \checkmark & 22.76 & 0.781 & 0.844 & 0.0026 & 0.0027 \\
HVS-YUV & \checkmark & \checkmark & \textbf{22.83} & \textbf{0.799} & \textbf{0.852} & 0.0026 & 0.0031 \\
\hline\hline
\end{tabular}\label{T3}
\end{table}

In addition, in order to achieve better visual performance, the proposed work applies structural loss to model training. In order to demonstrate its effectiveness, we analyze the experimental results on both the original task and the watermark task caused by applying different loss functions. In detail, $L_8$ indicates that both SSIM and MS-SSIM are used for model training. If $L_8$ is removed from Eq. (9) and Eq. (16), it means to skip the structural loss. By modifying $L_8$ as
\begin{equation*}
\frac{1}{|S_1|}\sum_{G(x)\in S_1}(1 - \text{SSIM}(G(x), T[G(x)])),
\end{equation*}
it means to only use the SSIM loss. By modifying $L_8$ as
\begin{equation*}
\frac{1}{|S_1|}\sum_{G(x)\in S_1}(1 - \text{MS-SSIM}(G(x), T[G(x)])),
\end{equation*}
it means to only use the MS-SSIM loss. Experimental results are shown in Table \ref{T3}. It can be inferred from Table \ref{T3} that using SSIM and MS-SSIM as part of the entire loss function indeed has the ability to further improve the visual quality of the marked images. Meanwhile, the BER differences are all low, meaning that the introduction of structural loss will not significantly impair the watermarking performance. In summary, the proposed two frameworks are suitable for practice.

\section{Conclusion and Discussion}
In this paper, we propose two watermarking frameworks based on HVS for generative models that output color images as the results. Unlike the previous methods that embed a watermark into the generated image directly, the proposed two frameworks select the more suitable channel of the image generated by the host network according to the characteristics of HVS for watermark embedding and watermark extraction. As a result, the marked image generated by the trained network has better quality compared with the previous art without impairing the watermark according to the reported experimental results. Though the structure and task of the host network are specified in our experiments, it is open for us to apply the proposed two frameworks to many other networks, indicating that this work has good universality.

On the other hand, as reported in the experimental section, we should admit that the binary watermarks may not be extracted perfectly. This is due to the reason that neural network learns knowledge from given data, which makes the neural network fall into the local optimum point that cannot perfectly model the mapping relationship between the marked image and the watermark. However, despite this, the BER can be kept very low, implying that by introducing error-correcting codes, the original watermark can be actually perfectly reconstructed.

From the viewpoint of robustness, even though augmenting the training data samples through mimicking attacks can improve the robustness of the model, real attacks are actually very complex and unpredictable. It is therefore necessary to further enhance the robustness of the model from the perspective of structural design and loss function design. It may also be very helpful by incorporating the interpretability theory of neural networks, which will be investigated in future.

\section*{Acknowledgement}
This work was supported by the CCF-Tencent Open Fund.

\end{document}